\documentclass[letterpaper, 10 pt, conference]{ieeeconf}  

\IEEEoverridecommandlockouts                              

\usepackage{graphicx} 
\graphicspath{{figures/}}
\usepackage{amsmath} 
\usepackage{amssymb}  
\usepackage[flushleft]{threeparttable}

\usepackage{color}
\usepackage{cite}
\makeatletter
\let\NAT@parse\undefined
\makeatother
\usepackage[colorlinks = true]{hyperref}
\usepackage{balance}

\title{\LARGE \bf
Early Turn-taking Prediction with Spiking Neural Networks \\ for Human Robot Collaboration
}

\author{Tian Zhou and Juan P. Wachs
\thanks{*Research supported partially by the Office of the Assistant Secretary of Defense for Health Affairs under Award No. W81XWH-14-1-0042, and partially by NPRP award (NPRP 6-449-2-181) from the Qatar National Research Fund (a member of The Qatar Foundation). The statements made herein are solely the responsibility of the authors. All authors are with the School of Industrial Engineering, Purdue University, West Lafayette, IN \texttt{\{zhou338, jpwachs\}{@}purdue.edu}}
}
\begin{document}

\maketitle
\thispagestyle{empty}
\pagestyle{empty}

\begin{abstract} Turn-taking is essential to the structure of human teamwork. Humans are typically aware of team members' intention to keep or relinquish their turn before a turn switch, where the responsibility of working on a shared task is shifted. Future co-robots are also expected to provide such competence. To that end, this paper proposes the Cognitive Turn-taking Model (CTTM), which leverages cognitive models (i.e., Spiking Neural Network) to achieve early turn-taking prediction. The CTTM framework can process multimodal human communication cues (both implicit and explicit) and predict human turn-taking intentions in an early stage. The proposed framework is tested on a simulated surgical procedure, where a robotic scrub nurse predicts the surgeon's turn-taking intention. It was found that the proposed CTTM framework outperforms the state-of-the-art turn-taking prediction algorithms by a large margin. It also outperforms humans when presented with partial observations of communication cues (i.e., less than 40\% of full actions). This early prediction capability enables robots to initiate turn-taking actions at an early stage, which facilitates collaboration and increases overall efficiency.

\end{abstract}

\section{INTRODUCTION} \label{sec:introduction}
Turn-taking provides humans the fundamental structure to organize and coordinate their collaborative actions in conversations \cite{sacks_simplest_1974}, collaborative task-solving \cite{inkpen_turn-taking_1997}, and shared haptic control \cite{chan_designing_2008}. During collaboration, each agent has the capability to comprehend the on-going task progress and their peer's multi-modal communication cues, in order to predict whether, when, and how to size the next turn. Having a fluent and natural turn-taking process can enhance the collaboration efficiency and strengthen the communication grounding among team members \cite{sebanz_joint_2006,marsh_social_2009}. Such coordinated turn-taking behaviors stand out clearly in high-risk and high-paced tasks like surgery in the Operating Room (OR). Surgeons and nurses perform fast, fluent, and precise turn-taking actions when exchanging surgical instruments. Therefore, the OR scenario was selected as the test-bed for investigating human robot turn-takings.  \par 

\begin{figure}[t]
\centering
\includegraphics[width=1.0\linewidth]{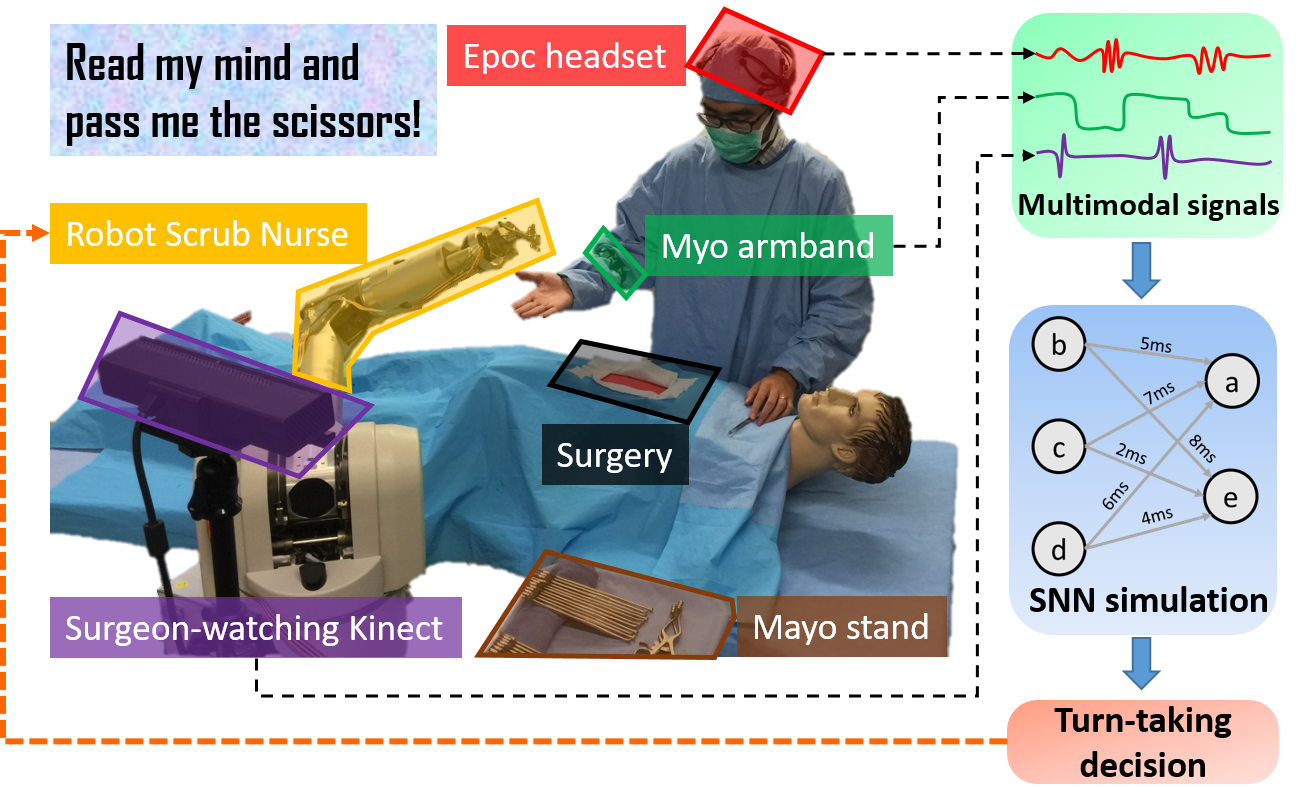}
\caption{Illustration of the robotic nurse system with turn-taking prediction. Various sensors (Kinect, Epoc, and Myo) are used to capture the surgeon's communication cues. The output of such network are precise inferences about the delivery of surgical instruments ahead of time.}
\label{fig:system}
\end{figure}

The same turn-taking behaviors observed during human-human interactions are expected in human-robot interactions. When robots are designed to collaborate with humans, they are expected to understand the human's turn-taking intentions, in order to determine when is a good time to engage in an interaction. When designing robotic nurses to collaborate with surgeons, they are expected to understand both the surgeon's implicit communication cues (e.g., change of body posture like leaning forward) and explicit cues (e.g., uttering the word ``scissors'') \cite{chao_timed_2016}. Figure \ref{fig:system} shows our human-robot system including the capability to understand both implicit and explicit communication cues for turn-taking. \par 

\begin{figure}[b]
\centering
\includegraphics[width=1.0\linewidth]{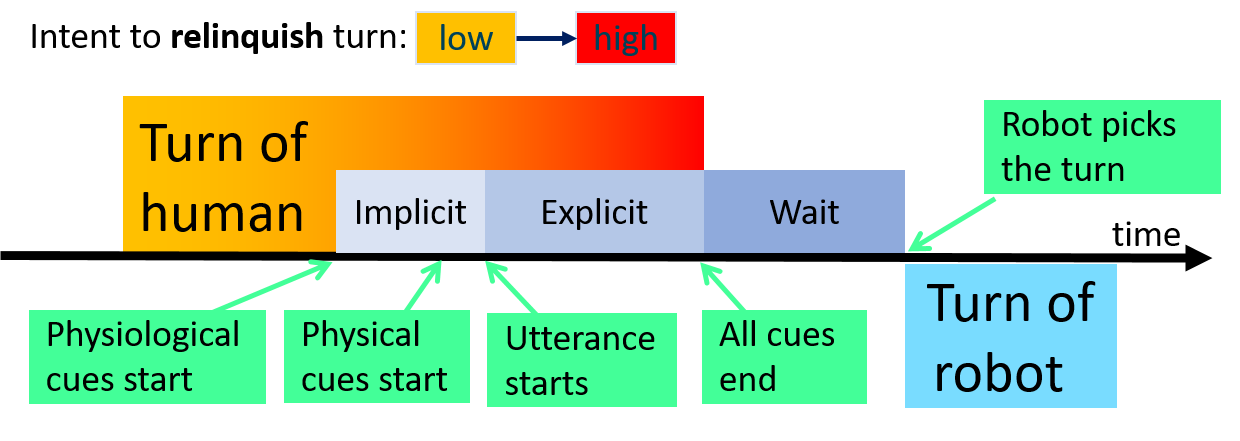}
\caption{Illustration of traditional human-robot turn transition process. The color transition from yellow to red indicates an increased level of human intent to relinquish the turn.}
\label{fig:turn}
\end{figure}

A typical process where humans delegate the turn to a robot is illustrated in Figure \ref{fig:turn}. As the human is approaching the end of her turn, she starts exhibiting implicit cues (i.e., physiological and physical cues), followed by explicit cues (i.e., utterance). The collection of these multimodal signals indicates her intention to relinquish the turn. Simultaneously, the robot needs to capture these subtle communication cues and predict the end of her turn. Early and accurate predictions of such turn-giving intentions allow robots to begin preparatory movements to facilitate the turn transition process.\par 

There has been research on developing algorithms to automatically recognize human's turn-taking intentions, in both conversations \cite{de_kok_multimodal_2009} and physical collaborations \cite{calisgan_identifying_2012}. However, current turn-taking recognition algorithms build on mathematically derived models such as Support Vector Machines \cite{arsikere_enhanced_2015}, Decision Trees \cite{loiselle_exploration_2005}, and Conditional Random Field \cite{de_kok_multimodal_2009}. Although a certain level of recognition accuracy is achieved, it still has not reached the level of human competence \cite{heeman_can_2015}. Moreover, a model derived computationally and mathematically cannot be explained and interpreted well by humans due to different reasoning processes. Therefore, there is a need of a cognitive turn-taking reasoning model that reaches the level of human competence and can be easily interpreted. \par 

To bridge this gap, the Cognitive Turn-Taking Model (CTTM) is proposed to simulate and predict human's turn-taking intentions. The CTTM is built upon the brain-inspired Spiking Neural Network (SNN). The SNN framework has plasticity structure \cite{maass_networks_1997} to model turn-taking processes and predicts human's turn-taking intentions based on multimodal observations. Compared to traditional neural networks, SNN has the advantage of modeling conduction delays of variable lengths. Therefore, it is suitable for time sensitive sequence modelling tasks such as gesture recognition \cite{botzheim_human_2012}, speech recognition \cite{loiselle_exploration_2005}, and seizure detection \cite{ghosh-dastidar_improved_2007}. However, there are still challenges in using SNN for multimodal turn-taking prediction such as how to map the raw signals into neurons, how to deal with multimodal fusion problems, and how to create descriptive features for early prediction. Those challenges are tackled by the proposed CTTM framework. To summarize, this paper makes the following three-fold contributions: 1) presents a formulation to rigorously define the human-robot collaborative task and the associated turn-taking events; 2) proposes the cognitive framework CTTM for early turn-taking prediction; 3) designs a multimodal interaction system between surgeons and robotic nurses.\par 

The paper is organized as the following: Section \ref{sec:related_work} presents the related work, followed by definitions of the collaborative task and turn-events in Section \ref{sec:definition}. The proposed CTTM framework is introduced in Section \ref{sec:cttm}, and the experimental setup and results are presented in Section \ref{sec:experiments}. Finally, Section \ref{sec:conclusion} summarizes the paper with conclusions and future work.

\section{RELATED WORK} \label{sec:related_work}
This section presents the related work about turn-taking analysis, in the context of conversational turn-taking and physical turn-taking. \par 

In conversational turn-taking analysis, linguistic cues such as pause duration \cite{schlangen_reaction_2006} and pitch levels \cite{ward_dialog_2010} have been found to play a key role in turn-taking regulation. The study of de Ruiter el al. \cite{ruiter_projecting_2006} revealed that syntax and semantics cues are also important in projecting the end of a speaker's turn. Non-verbal behaviors such as gaze and posture shifts  have been investigated and found to be relevant to turn regulations \cite{padilha_nonverbal_2003}. \par 

Physical turn-taking refers to the process where a hybrid human-robot team take turns on a physical task. Turn-taking has been studied in robotics and HRI to regulate shared resources among team members. Those resources include time (i.e., only one person can work on the shared task at a time) and space (i.e., only one person can access the working space at a time). In the context of human robot interaction, the types and usage frequencies of various implicit communication cues have been studied in a robot-assisted assembly task \cite{calisgan_identifying_2012}. The timing in multimodal turn-taking (i.e., speech, gaze, and gesture) was investigated through a collaborative Towers of Hanoi challenge accomplished by a hybrid human robot team \cite{chao_timed_2012}. Nonverbal cues for timing coordination in physical turn-taking were studied in manufacturing \cite{hart_gesture_2015}. Such research focuses on turn-taking process modelling and the robot control, while neglecting predictions of the turn-taking timing and intention. 

\section{COLLABORATIVE TASK AND TURN-EVENTS DEFINITION} \label{sec:definition}
This section introduces the definitions of collaborative task and associated turn-events to formulate the early turn-taking prediction problem. \par 

Consider a human agent $H$ working with a robot agent $R$ on a collaborative task $\mathcal{W}$. $H$ and $R$ conduct and alternate through a sequence of subtasks $w_k^A$, where subscript $k$ indicates the subtask index and superscript $A$ indicates the agent responsible for this subtask (i.e., $A \in \{H,R\}$). The subtask is defined as $w_k^A\triangleq(g_k,z_k^b,z_k^f)$, where $g_k\in G$ is the action label, $z_k^b$ is the beginning time, and $z_k^f$ is the finishing time of the subtask. $G$ is the set containing all the action labels, such as delivering, retrieving, and exchanging tools between agents. The subtask $w_k^A$ is treated as the ``atomic'' component in the definition since a turn only happens during the transition of two subtasks. \par 

\begin{figure}[b]
\centering
\includegraphics[width=1.0\linewidth]{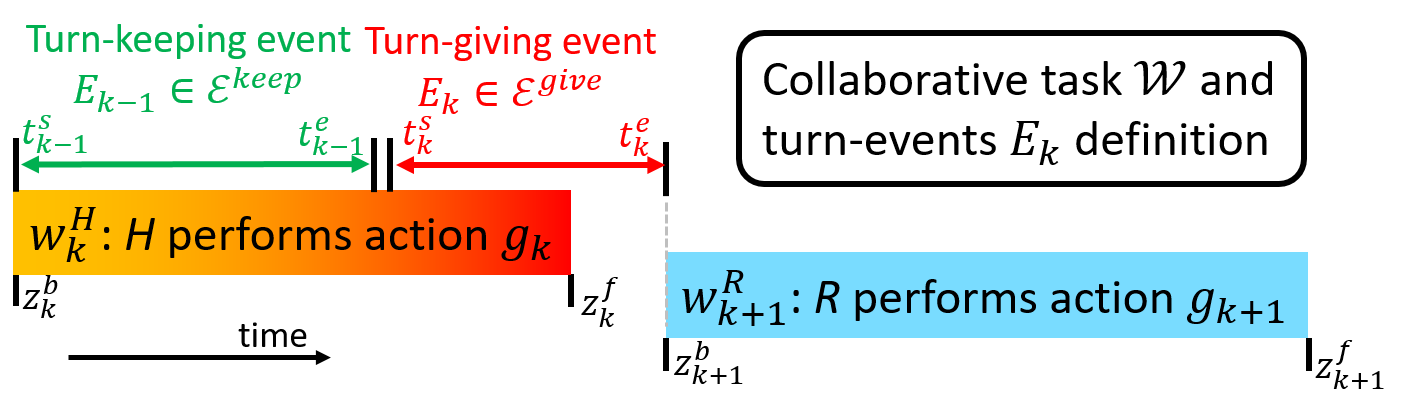}
\caption{Illustration of the definition of collaborative task $W=\{w_k^A\}$ and turn-events $E_k \in \{\mathcal{E}^{give},\mathcal{E}^{keep}\}$.}
\label{fig:def}
\end{figure}

As the human $H$ conducts subtask $w_k^H$ and time goes on from $z_k^b$ to $z_k^f$, she gets closer in finishing the subtask and expresses an increased level of intention to give out the turn. Since this paper focuses on enabling robotic assistants to take turns from humans, only the turn transitions from $H$ to $R$ are considered. Each turn transition from $w_k^H$ to $w_{k+1}^R$ defines a turn-event $E_k$, in which the human is showing an intention to give out the turn (denoted as $\mathcal{E}^{give}$). On the other hand, for the majority of subtask $w_k^H$, the human is focusing on the current action and shows no intention to relinquish the turn. This period implicitly defines a turn-event $E_k$, in which human intends to keep the turn (denoted as $\mathcal{E}^{keep}$). Each turn-event $E_k \in \{\mathcal{E}^{give},\mathcal{E}^{keep}\}$ spans a time window $[t_k^s,t_k^e]$, which are the starting and ending time for this turn-event. The collaborative task $\mathcal{W}$, subtask $w_k^A$, and turn-events are illustrated in Figure \ref{fig:def}. \par 

As the human is conducting subtask $w_k^H$, she is monitored through $M$ sensor channels which include physiological, physical, and neurological signals. The captured signal at time $t$ is denoted as $\vec{s}(t) \in \mathbb{R} ^ {1 \times M}$. Given a turn-event $E_k$ which spans time $[t_k^s,t_k^e]$, the captured signals within this time window are stacked together to form a matrix representation $X_k \triangleq [\vec{s} (t_k^s:t_k^e) ] \in \mathbb{R}^{L_k \times M}$, where $L_k$ is the discrete event length (i.e., $L_k=t_k^e-t_k^s$). For each stacked segment $X_k$, a label $y_k\in\{0,1\}$ is assigned to indicate either turn-giving (i.e., $y_k=1$ when $E_k \in \mathcal{E}^{give}$) or turn-keeping (i.e., $y_k=0$ when $E_k \in \mathcal{E}^{keep}$). The turn-taking prediction algorithm $\phi(\cdot)$ then returns an estimate $\hat{y}_k = \phi(X_k) \in \{0,1\}$. Moreover, the turn-taking intention should be predicted in an early stage given only partial observations (i.e., with only implicit cues before explicit cues even start). This way, the human's intent to relinquish the turn can be recognized early and the robot can start moving early to facilitate the transition. Therefore, given partial observation $X_k^ \tau \in \mathbb{R}^{(\tau L_k) \times M}$ for only the beginning $\tau$ fraction ($0<\tau\leq1$), an early decision is made according to $\hat{y}_k^\tau = \phi(X_k^\tau) \in \{0,1\}$. The resultant dataset $D^\tau=\{(X_k^\tau,y_k,\hat{y}_k^\tau)  | X_k^\tau \in \mathbb{R}^{(\tau L_k) \times M},y_k \in \{0,1\}, \hat{y}_k^
\tau = \phi(X_k^\tau ) \in \{0,1\}  \}$ is then used for the following turn-taking analysis.

\section{COGNITIVE TURN-TAKING MODELS (CTTM)} \label{sec:cttm}
This section presents the CTTM framework: Section \ref{sec:snn_structure} discusses the network structure, Section \ref{sec:snn_mapping} presents the neuron mapping methods, and Section \ref{sec:snn_training} discusses the various aspects of SNN training. 

\subsection{SNN Structure} \label{sec:snn_structure}
Conventional neural networks enforce the same conducting delay between consecutive layers, thus all the neurons of the same layer can only fire at the same time. This rigid structure is insufficient when modeling multimodal temporal sequences [20]. SNN, however, can model the variability of axonal conduction delays between neurons. Firings will take different amounts of time to traverse around the network based on the different conduction delays. This way, the asynchronous effect of multimodal signals can be modeled. 


The adopted SNN structure and parameters are similar to the work of Izhikevich \cite{izhikevich_polychronization:_2006}. The network has 250 neurons, with 200 excitatory neurons (i.e., can be stimulated) and 50 inhibitory neurons (i.e., cannot be stimulated). Each excitatory and inhibitory neuron has 25 post synapses, connecting to 25 other neurons following a uniform distribution. Each synapse has a conduction delay in the range of [1, 20] ms, following a uniform distribution. The weights of the synaptic connections are initialized to be +6 for all post synapses after excitatory neurons, and -5 for all post synapses after inhibitory neurons. Those weights represent how strong the synaptic connection is between two neurons, and are updated based on the Spike Timing-Dependent Plasticity (STDP) rule \cite{sjostrom_spike-timing_2010}. The maximum weight for each synaptic connection is set to be 10. \par

Each neuron model is depicted by a set of formulas, as given in (\ref{eq:snn}). This neuron model, together with the constant parameters are the same with the simple spiking model proposed in \cite{izhikevich_which_2004}:
\begin{align}
\begin{split}
 \frac{dv}{dt} &= 0.04v^2 + 5v + 140 - u + I \\
 \frac{du}{dt} &= a(bv-u) 
\end{split}
\label{eq:snn}
\end{align}

with the auxiliary after-spike resetting:
\begin{equation}
\text{if} \, v \geq \, +30 \,mV, \,\text{then} 
	\begin{cases}
    v \leftarrow c\\
    u \leftarrow u+d
    \end{cases}
\label{eq:snn2}
\end{equation}

\begin{figure}[thpb]
\centering
\includegraphics[width=1.0\linewidth]{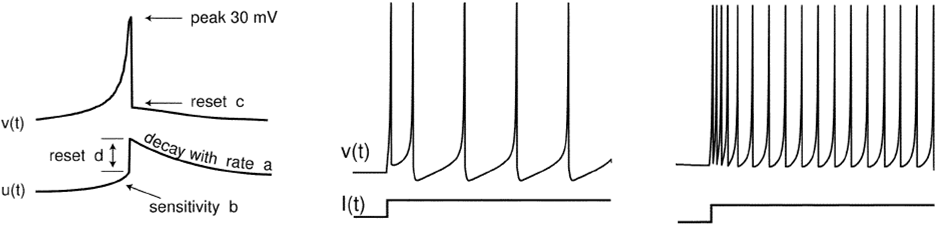}
\caption{(left) dynamics of the simple spiking model. (middle) voltage response of regular spiking (RS) neuron to a step DC-current $I$=10 mA. (right) voltage response of low-threshold spiking (LTS) neuron. Time resolution is 0.1ms. Electronic version of the figure and reproduction permissions are freely available at www.izhikevich.com.}
\label{fig:snn}
\end{figure}

The variable $v$ represents membrane potential of the neuron, and $u$ represents a membrane recovery variable which provides negative feedback to $v$. Variable $I$ is the input DC current to this neuron, which was set to 20 mA when this neuron is stimulated by the input multimodal data. As illustrated in Figure \ref{fig:snn}, for recovery variable $u$, $a$ represents the time scale, $b$ represents the sensitivity, and $d$ represents after-spike reset increment. For membrane potential variable $v$, $c$ represents the after-spike reset value. The excitatory neurons are configured to exhibit regular spiking (RS) firing patterns with parameters $(a,b,c,d)=(0.02,0.2,-65,8)$, and the inhibitory neurons are configured to exhibit low-threshold spiking (LTS) firing patterns with parameters $(a,b,c,d)=(0.02,0.25,-65,2)$. Those parameters were chosen since they performed the best out of all excitatory-inhibitory neuron combinations proposed by \cite{izhikevich_which_2004} in our case.

\subsection{Neuron Mapping} \label{sec:snn_mapping}
To model human turn-taking behaviors, it is necessary to map the input multimodal data to the SNN neurons. Previous research has attempted to map the discrete input data into the neurons on a one-to-one basis. For example, one image pixel corresponds to one neuron \cite{rekabdar_scale_2015} or one level of orientation maps to 5 randomly chosen neurons \cite{beyeler_categorization_2013}. However, mapping multimodal continuous-valued signals into SNN neurons requires a different approach since the network dimension grows exponentially to encode all possible combinations of continuous inputs. A small example of 5 discrete levels and 10 multimodal channels would lead to $10^8$ neurons, which is intractable. This problem is solved by resorting to automatic channel quantization and decision-level fusion methods.\par 

First, each of the $M$ channels was quantized into $V$ discrete levels. The 1-percentile threshold ($r_1$) and the 99-percentile threshold ($r_{99}$) are used as fences to remove outliers. The $V$ bins were divided within the $1\% \sim 99\%$ range to encode the continuous sensor signals. Given a sensor reading value $s$, it was quantized to level $q\, (q=0,1,...,V-1)$, following $q=int(\frac{s-r_1}{r_{99}-r_1}*V)$. The quantization process is applied to each of the $M$ channels of $X_k \in \mathbb{R}^{L_k \times M}$. The result is $\tilde{X}_k \in \mathbb{Q}_V^{L_k \times M}$, where $\mathbb{Q}_V$ represents the quantized space with $V$ levels. Now the partial observation is $\tilde{X}_k^\tau \in \mathbb{Q}_V^{(\tau L_k) \times M}$. For each quantized level $q$, 5 excitatory neurons in the SNN were randomly allocated following a uniform distribution to represent it. When level $q$ is active, all its 5 corresponding neurons are stimulated one by one at 1 ms intervals. The DC current of 20 mA provides the stimulation to variable $I$ in (\ref{eq:snn}). The value of $V$ was set to be 40, to reach a total of $40*5=200$ excitatory neurons so that each excitatory neuron is mapped to a discrete level. \par 

To encode the multimodal inputs, one SNN is constructed for each of the M channels. The outcomes from each SNN are combined in the end using a decision-level fusion. This approach follows the human brain mechanism (i.e., vision and hearing are independently processed and are fused in later cognitive decision-making stage).

\subsection{SNN Training} \label{sec:snn_training}
Once the mappings between input data $\tilde{X}_k^\tau$ and SNN neurons are established, the network needs to be trained. The SNN training includes three stages, where the first stage trains the SNN network weights by feeding unsupervised training data repeatedly into the network. Each training observation activates corresponding neurons in sequence, and the network weights are updated following the STDP rule. The second stage consists of soliciting network responses for different turn-taking classes, and the third stage focuses on constructing features from SNN responses.

\subsubsection{SNN Synapse Weight Training} 
This training phase aims to tune the synaptic weights for SNN so that it can properly encode the input spatio-temporal signals. For that purpose, the STDP training is used. Under STDP, the synaptic weights are updated based on timing differences of the neural firings \cite{izhikevich_simple_2003}. The synaptic weights between those neurons which always fire together are strengthened. More specifically, the weight of synaptic connection from pre- to postsynaptic neuron is increased if the post-neuron fires after the presynaptic spike and is decreased otherwise. Parameters for STDP training are set based on \cite{izhikevich_polychronization:_2006}. During this training stage, each quantized training data $\tilde{X}_k \in \mathbb{Q}_V^{L_k \times M}$ is mapped to its corresponding neurons in the SNN, and the synaptic weights are updated in each 1ms interval based on the STDP rules. The time allocated to simulating each $\tilde{X}_k$ is 250ms. Since the input data length $L_k$ is less than 40, it requires less than 200ms to stimulate the network. Then the network continues to run for 50ms without being provided any input, to allow the spike trains to propagate the network. Patterns $\tilde{X_k}$ for both turn-taking classes ($y_k \in \{0,1\}$) are presented to the SNN network during this training phase, following a random repeated order. The network was simulated for 250s, which includes a total of 1000 training inputs $\tilde{X}_k$, each of which takes 250ms to simulate. After the simulation, the synaptic weights in the network converge into a steady state. 

\subsubsection{Signature Firing Patterns Training}
This training phase aims to generate the stereotypical network responses for different types of inputs $\tilde{X}_k$. As mentioned earlier, one SNN network was constructed for each channel of information (i.e., one column of $\tilde{X}_k \in \mathbb{Q}_V^{L_k \times M}$, denoted as $\tilde{X}_{ki}$ for column $i$). Therefore, there are $M$ SNN constructed in total, forming a SNN group and is denoted as $\mathcal{S}=\{S_i\}, i=1,...,M$. Given input $\tilde{X}_k$, its response to this SNN group is denoted as $\mathcal{G}_k=\mathcal{S}(\tilde{X}_k)$. $\mathcal{G}_k$ consists of $M$ individual responses ($G_{ki}$) for each of the SNN networks, i.e., $\mathcal{G}_k \triangleq \{G_{ki} \}, i=1,...,M$ where responses $G_{ki} \triangleq S_i(\tilde{X}_{ki})$. $G_{ki}$ denotes the firing maps (which neurons fired at what time) when input $\tilde{X}_{ki}$ is presented to the model $S_i$. As mentioned earlier, given an input $\tilde{X}_{ki}$, the network was simulated for $T$ milliseconds and each millisecond forms an atomic time unit. There are in total $N$ neurons in the network which can be potentially fired. Therefore, $G_{ki}$ is formed as a $N$ by $T$ Boolean matrix (i.e., $G_{ki}\in\mathbb{B}^{N\times T}$), where a value of 1 at cell $(n,t)$ indicates that neuron $n \,(1\leq n \leq N)$ fired at time $t \,(1 \leq t \leq T)$, i.e.,:
\begin{equation}
G_{ki}(n,t) = \begin{cases}
    1 & \text{neuron n fired at time t} \\
    0 & \text{neuron n did not fire at time t}
    \end{cases}
\label{eq:snn3}
\end{equation}

The firing map $\mathcal{G}_k = \{G_{ki}\}$ forms a compact and rich representation of the original signal $X_k$, and is used to predict the turn-taking type ($\hat{y}_k \in \{0,1\}$). When given partial observation $X_k^\tau$, its discretized version $\tilde{X}_k^\tau$ is fed into the SNN group $\mathcal{S}$, generating a partial response $\mathcal{G}_k^\tau = \mathcal{S}(\tilde{X}_k^\tau)$. This response is used to predict its turn-taking type $\tilde{y}_k^\tau$.

\subsubsection{Descriptive Feature Construction}
\begin{figure}[t]
\centering
\includegraphics[width=1.0\linewidth]{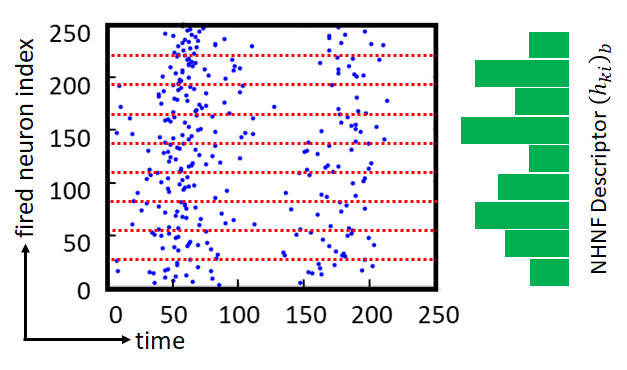}
\caption{Illustration of the NHNF descriptor.}
\label{fig:nhnf}
\end{figure}

This training phase focuses on creating descriptive features for the SNN network response $G_{ki}$. Although $G_{ki}$ can be directly used for classification purposes, it would cost unnecessary computational time and memory usage, as $G_{ki}$ is a large sparse matrix with many zeros. To solve that problem, we proposed the Normalized Histogram of Neuron Firings (NHNF) descriptor to compactly represent $G_{ki}$. Specifically, the total number of neurons (i.e. $N$) are evenly divided into $B$ bins, where bin $b\, (b=0,…,B-1)$ covers neurons indexes in the range $[bN/B,(b+1)N/B]$. During a simulation of $T$ ms, the total number of firings for neurons within bin $b$ is counted and then divided by the simulation duration $T$ to generate the descriptor $(h_{ki})_b$ for sample $X_k$ and feature $i$:
\begin{equation}
(h_{ki})_b = \frac{1}{T} \sum_{t=1}^T{\sum_{n=bN/B}^{(b+1)N/B}{G_{ki}(n,t)}}
\label{eq:NHNF}
\end{equation}

for $i=1,...,M;k=1,...,K;b=0,...,B-1$. Dividing the histogram by the simulation duration $T$ can make this descriptor duration-invariant and thus can work robustly for variable signal lengths. An illustration of the NHNF descriptor for a sample firing output is given in Figure \ref{fig:nhnf}. In the figure, the red dotted lines indicate the boundary of bins, and the green bars represent the histogram values. \par 

Since there are $M$ channels of information in total, $M$ sets of histograms $(h_{ki})_b$ are generated for a given sample $X_k$. The histograms $(h_{ki} )_b$ for each bin ($b$) and each channel ($i$) are concatenated together to form the final feature descriptor for input $X_k$, denoted as $H_k$. Then $H_k$ is used to predict the turn-event type $\hat{y}_k$. When only partial responses $\mathcal{G}_k^\tau$ are available, the NHNF descriptors were extracted from it (denoted as $H_k^\tau$) and used to predict the turn-event type $\hat{y}_k^\tau$.

\section{EXPERIMENTS} \label{sec:experiments}
The proposed CTTM framework was tested in a robotic scrub nurse scenario, where the surgeon's turn-taking intentions must be predicted ahead of time. This section discusses the relevant aspects in the experiment setup, including surgical task setup (Section \ref{sec:task_setup}), human sensing and signal processing (Section \ref{sec:sensing}), the performance (Section \ref{sec:cttm_performance}), and finally the network visualization (Section \ref{sec:visualization}).

\subsection{Surgical Task Setup} \label{sec:task_setup}
A simulation platform for surgical operations was used to capture turn-taking cues of surgeons, as shown in Figure \ref{fig:system}. The platform consists of a patient simulator and a set of surgical instruments to conduct a mock abdominal incision and closure task \cite{martyak_abdominal_1976}. In this collaborative task, the surgeon and nurse collaborate by exchanging surgical instruments. The surgeon performs operations while the nurse searches, prepares, and delivers the requested surgical instrument. \par 

Participants were recruited to perform a mock surgical task. Twelve participants were recruited from a large academic institution, with the age range of 20 to 31 years (M=25.7, SD=2.93). After inform consent was given (IRB protocol 1305013664), participants completed a training session. Surgical instruments were introduced by repeated recitation of their names, and a training video of step-by-step instructions of the mock abdominal incision and closure task (10 mins) was given. After the video, the participant attempted a ``warm-up'' trial. Each participant repeated the surgical task 5 times to reach performance proficiency. Although a surgeon population was not used, the training sessions and repeated trials led to high face-fidelity data. This dataset is realistic enough to validate the early turn-taking prediction capability. \par 

Each execution of the surgical task included in average 14 surgical instrument requests. The surgical request actions were annotated as turn-giving events ($\mathcal{E}^{give}$), and the surgical operation actions were annotated as turn-keeping events ($\mathcal{E}^{keep}$). Two annotators were presented with the recorded videos of the surgical tasks. The annotators independently determined the starting and ending time for each turn-event as well as its type. The main annotator labeled the entire dataset while the second annotator labeled a random 10\% of selected segments. An inter-rater reliability of Cohen's $\kappa=0.95$ \cite{cohen_coefficient_1960} was found, indicating high agreement between two sets of annotations. Overall, 846 turn-giving events ($y_k=1$) and 1305 turn-keeping events ($y_k=0$) are generated for the following turn-taking analysis.

\subsection{Multimodal Human Sensing and Signal Processing} \label{sec:sensing}
\begin{figure}[thpb]
\centering
\includegraphics[width=1.0\linewidth]{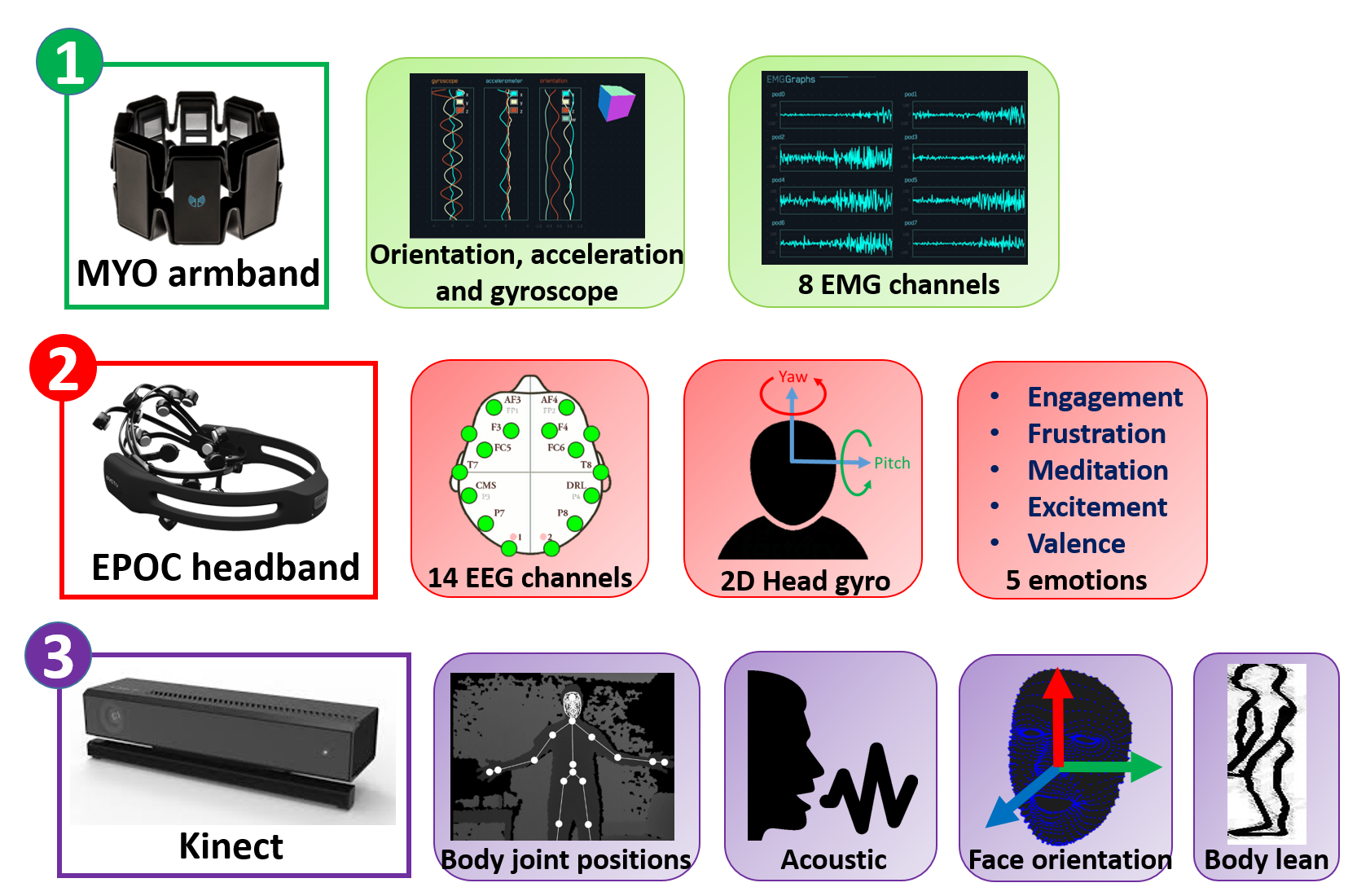}
\caption{Multimodal human sensing. The three sensors were used simultaneously to capture human's multimodal signals.}
\label{fig:sensor}
\end{figure}

The participant in the surgeon role had her communication cues collected during the surgical operation for turn-taking analysis. Three sensors were used to capture the multimodal communication cues: the Myo armband, Epoc headset, and Kinect sensor. Each sensor captured multiple channels of information from the human, as illustrated in Figure \ref{fig:sensor}. There were in total $M (M=50$) channels of raw signals, which were  synchronized at a frequency of 20 Hz. 

Preprocessing techniques were used to smooth and normalize the raw multimodal signals. Each of the $M$ channels of information was smoothed with Exponentially Weighted Moving Average approach with an empirical weight of 0.2 \cite{lucas_exponentially_1990}. Then, each of the $M$ channels was normalized to zero mean and unit variance.\par 

The feature construction and selection algorithm proposed by \cite{morency_context-based_2008} was adopted. Each channel of the normalized signals was first convolved with a filter bank containing 6 filters, i.e., identity transformation, Sobel operator, Canny edge detector, Laplacian of Gaussian detector, and two Gabor filters. Then the correlation between each encoding with the turn-event labels was calculated through $\chi^2$ test of independence. The $m$ features of the largest test statistic values were retained as the final feature set to form the initial representation $X_k$. In this experiment, the value of $m$ was set to 10 empirically. The selected top 10 features are shown in TABLE \ref{tab:feature}. 

\begin{table}[htb]
\caption{Selected Top Features}
\label{tab:feature}
\vspace{-3mm}
\begin{center}
\begin{tabular}{lll}
\hline
Rank & Feature name + Filter name & $\chi^2$ \\
\hline
1 & Epoc.gyro\_y + identity & 1479.2 \\ 
2 & Epoc.gyro\_y + gabor1 & 1456.6 \\
3 & Epoc.gyro\_y + gabor2 & 1430.9 \\
4 & Kinect.audioConfidence + gabor1 & 1424.7 \\
5 & Kinect.audioConfidence + identity & 1408.5 \\
6 & Kinect.audioConfidence + gabor2 & 1388.0 \\
7 & Myo.orientation\_x + gabor1 & 990.3 \\
8 & Myo.orientation\_x + gabor2 & 975.9 \\
9 & Myo.acceleration\_y + gabor1 & 975.1 \\
10 & Myo.acceleration\_y + gabor1 & 971.1 \\
\hline
\end{tabular}
\end{center}
\end{table}

\subsection{CTTM Performance} \label{sec:cttm_performance}
To evaluate the performance of CTTM in predicting surgeons' turn-taking intentions, computational experiments were conducted. The experiment setup followed the leave-one-subject-out (loso) cross validation. In each fold, the data from 11 subjects was used for training and data from the last subject was separated for testing. Such evaluation scheme can evaluate the algorithm's generalization capability on unseen subjects. For accuracy measurement between prediction result  $\hat{y}_k \in \{0,1\}$ and ground truth $y_k\in\{0,1\}$, the $F_1$ score for turn-giving class was calculated (i.e., harmonic mean of precision and recall). \par 

The CTTM can recognize the type of the turn-event given only partial observation $X_k^\tau \in \mathbb{R}^{(\tau L_k) \times M}$. An early decision was then made according to $\hat{y}_k^\tau = \phi(X_k^\tau) \in \{0,1\}$. To evaluate the algorithm's early prediction performance, the $F_1(\tau)$ value for $\tau = 0.1, 0.2, ..., 1.0$ was calculated and presented. The NHNF descriptor $H_k^\tau$ was extracted from the beginning $\tau$ fraction of input (i.e., $X_k^\tau$), with 50 bins for the histogram (i.e., $B=50$). Then the descriptor $H_k^\tau$ was normalized so that each dimension had zero mean and unit variance. The normalized feature was then fed into a SVM classifier, which gave the prediction of turn-taking event type $\hat{y}_k^\tau$. The hyper-parameters for the SVM was set based on a 5-fold within-group grid search process. \par 

Four benchmark algorithms were implemented to compare and evaluate the proposed framework, as explained below.  \par 

The \underline{\smash{first benchmark, Dynamic Time Warping (DTW)}}, is a traditional time-series modelling algorithm. It has been successfully applied to speech recognition \cite{abdulla_cross-words_2003} and early gesture recognition \cite{mori_early_2006} but never to turn-taking tasks. The multi-dimensional DTW proposed in \cite{ten_holt_multi-dimensional_2007} was used with 1-norm distance measurement for two multi-dimensional signals. The k-nearest-neighbor classification scheme was applied (with 20 templates for each class). The DTW distance between the two feature vectors was calculated (i.e., $X_u$ and $X_k$) and the label of the closer sample was used as the predicted label. \par 

The \underline{\smash{second benchmark (Ishii)}} is a turn-taking prediction algorithm applied to human conservation \cite{ishii_predicting_2015}. Even though this framework targets a different turn-taking application, it can still be adopted into this case with minor modifications, as described below. In Ishii's framework, each signal channel from $X_k$ was normalized into the range of $[\mu-\sigma, \mu +\sigma]$ and then three descriptive features were extracted (i.e., average movement, average amplitude, and average frequency of movement) for each channel of signal. The SVM algorithm was then used for classification, with hyper-parameters selected based on 5-fold grid search. \par 

The \underline{\smash{third benchmark (SNN-PNG)}} is a SNN-type algorithm \cite{rekabdar_using_2017} and is the closet algorithm to the proposed approach. The major difference is that the SNN-PNG framework is a template-based technique and uses PNG as features. In \cite{rekabdar_using_2017}, PNG extracted from $G_{ki}$ were used as features, and the Jaccard similarity and Longest Common Sequence (LCS) algorithm are used to measure the distance between the unknown pattern and the training templates. A nearest-neighbor approach was used for classification purposes, with 20 templates for each class. \par 

The \underline{\smash{fourth benchmark (Human)}} reflects human performance. We used a ``button-press'' paradigm \cite{magyari_early_2014}, where recorded videos of the surgical operation were played back to participants and paused at random times. At every pause, the participant was asked what he thinks that the surgeon wants to do (keep or relinquish the turn). The participants in this experiment used a cross-participant setting for data annotation (no self-annotation). \par 

\begin{figure}[!b]
\centering
\includegraphics[width=1.0\linewidth]{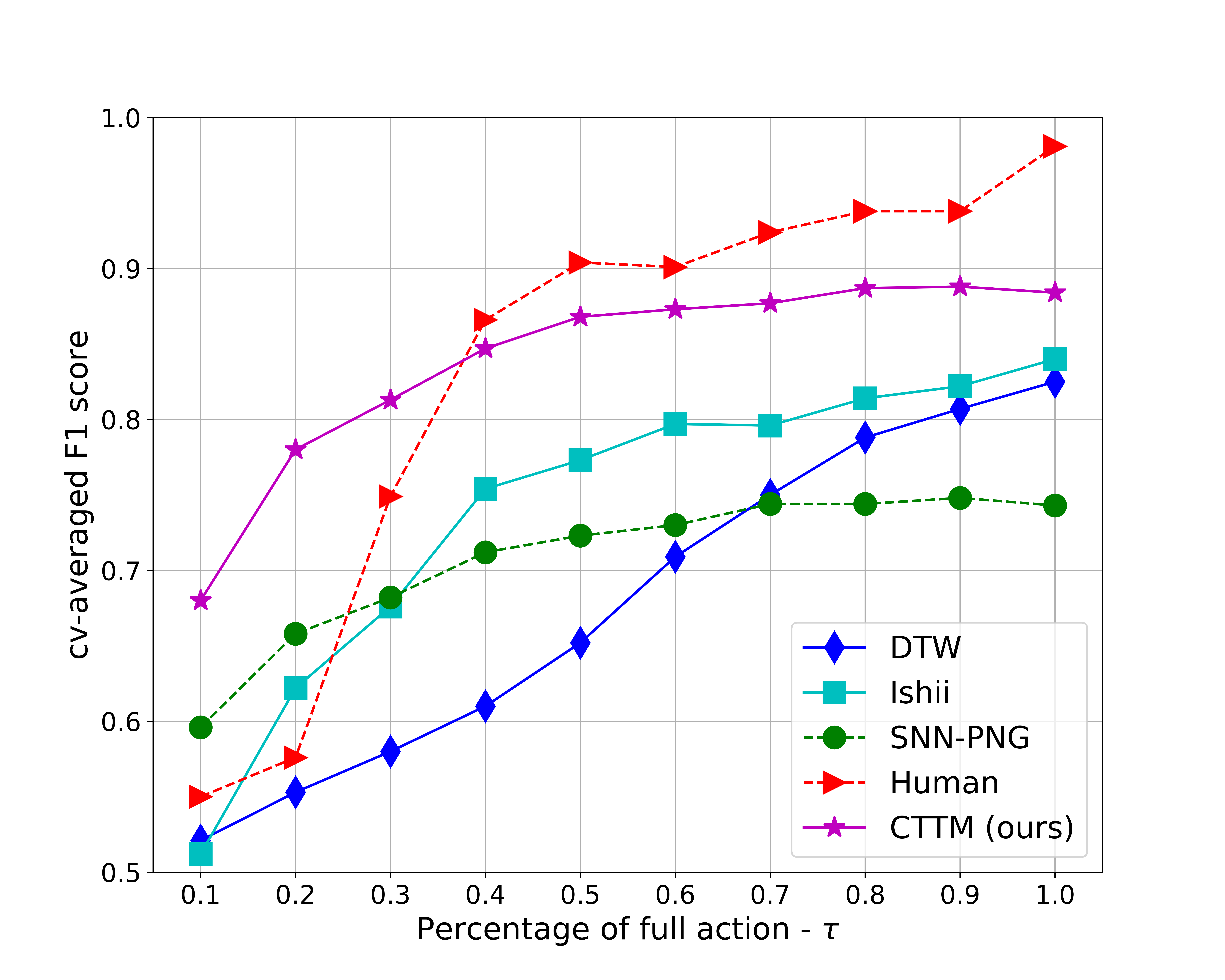}
\caption{Performance of proposed CTTM framework compared to the benchmark algorithms.}
\label{fig:f1_curve}
\end{figure}

The performances of the proposed CTTM framework with the four benchmarks are shown in Figure \ref{fig:f1_curve}. The CTTM framework greatly outperforms all the computational benchmark methods. Additionally, the CTTM performance surpasses human performance when little action is given (i.e., when $\tau<0.4$). After providing more observations, the human performance is better than CTTM, with an average F1 score margin of $0.05$. This prediction behavior can allow inference of human's turn-taking intentions in an early stage, since only a few anchor neurons are required to fire to generate a sequence of signals that traverse through the network, forming a stereotypical response \cite{rekabdar_scale_2015}. Similar observations have been reported in hand digit recognition tasks \cite{rekabdar_using_2017,rekabdar_scale_2015} and gesture recognition \cite{botzheim_human_2012} with SNN. 

\subsection{Visualization of Fired Neurons} \label{sec:visualization}
\begin{figure}[thpb]
\centering
\includegraphics[width=1.0\linewidth]{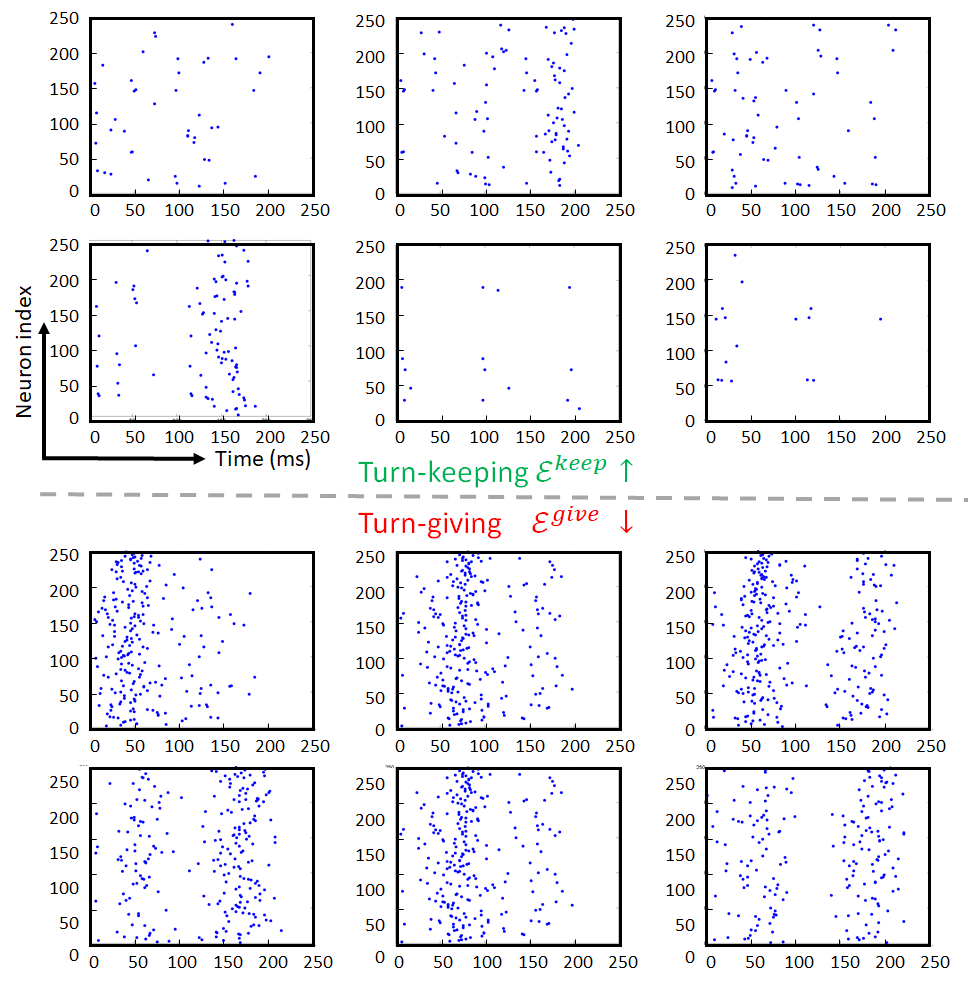}
\caption{Comparison of SNN responses to different classes of input. The SNN firing map is shown for the turn-keeping inputs (top two rows) and turn-giving inputs (bottom two rows). Each simulation lasts 250ms (x-axis) and there are 250 neurons (y-axis). Blue dot indicates neuron firing.}
\label{fig:firing}
\end{figure}

Visualizing the SNN responses allows better understanding of the patterns learned by the model. Figure \ref{fig:firing} shows 6 neurons firing maps (i.e., $G_{ki}$) for each class of input. The SNN model for the first feature was selected here for visualization. The responses to turn-keeping inputs ($\mathcal{E}^{keep}$) are on the top two rows, and the responses to the turn-giving inputs ($\mathcal{E}^{give}$) are at the bottom two rows. Visual inspection reveals that the SNN responds differently to $\mathcal{E}^{keep}$ and $\mathcal{E}^{give}$ inputs. The $\mathcal{E}^{give}$ inputs in general can fire more neurons in the trained SNN network compared to $\mathcal{E}^{keep}$ inputs. This could mean that humans exhibit a coherent pattern when relinquishing their turn. The neurons in the CTTM framework fire in the presence of such pattern. On the other hand, the intention to keep the current turn (i.e., focusing on operation) can be diverse (since the operations can be very diverse) and cannot trigger enough firing. Additionally, responses in $\mathcal{E}^{give}$ generally have a column-wise structure (either one column or two columns). This structure is generated when a group of neurons fire together in a time-locked pattern, forming a PNG as a signature of early turn-taking intent. 

\section{CONCLUSIONS} \label{sec:conclusion}
In human robot interaction, turn-taking capability is critical to enable robots to interact with humans seamlessly, naturally, and efficiently. However, current turn-taking algorithms cannot help accomplish early prediction. To bridge this gap, this paper proposes the CTTM, which leverages cognitive models for early turn-taking prediction. More specifically, this model is capable of reasoning human turn-taking intentions, based on the neurons firing patterns in the SNN. The CTTM framework can reason about the multimodal human communication cues (both implicit and explicit) and predict a person's intention of keeping or relinquishing the turn in an early stage. Such prediction can then be used to control robot actions. \par 

The proposed CTTM framework was implemented in a surgical context, where a robotic scrub nurse predicted the surgeon's turn-taking intentions to determine when to deliver surgical instruments. The algorithm's turn-taking prediction performance is evaluated based on a dataset, acquired through a simulated surgical procedure. The proposed CTTM framework outperformed computational state-of-the-art algorithms and can surpass human performance when only partial observation is available (i.e., less than 40\% of full action). Specifically, the proposed framework achieves a $F_1$ score of $0.68$ when only 10\% of full action is presented and a $F_1$ score of $0.87$ at 50\% presentation. Such early prediction capability is partially due to the implemented cognitive models (i.e., SNN) for early prediction. Such behavior would enable co-robots to work in a hybrid environment side by side with humans. \par 

Future work includes 1) including more contextual information to improve early prediction capability (e.g., current task progress); 2) validating the framework in real surgeries; 3) transfer the CTTM framework to other scenarios, such as robot-assisted manufacturing.

\balance
\bibliographystyle{ieeetr}
\bibliography{myZoteroBib}

\end{document}